**RESEARCH**  **Open Access**

# Blind Federated Learning without initial model

Jose L. Salmeron[1,2] and Irina Arévalo[1*]

*Correspondence:
irina.arevalo@cunef.edu

[1] CUNEF Universidad, Madrid, Spain
[2] Universidad Autónoma de Chile, Santiago, Chile

**Abstract**

Federated learning is an emerging machine learning approach that allows the construction of a model between several participants who hold their own private data. This method is secure and privacy-preserving, suitable for training a machine learning model using sensitive data from different sources, such as hospitals. In this paper, the authors propose two innovative methodologies for Particle Swarm Optimisation-based federated learning of Fuzzy Cognitive Maps in a privacy-preserving way. In addition, one relevant contribution this research includes is the lack of an initial model in the federated learning process, making it effectively blind. This proposal is tested with several open datasets, improving both accuracy and precision.

**Keywords:** Federated learning, Privacy-preserving machine learning, Fuzzy Cognitive Maps

## Introduction

Federated learning is an emerging approach to enable privacy-preserving machine learning by sharing local models instead of the data itself. Therefore, it is a method for training machine learning models in a distributed way, and it can be used for both classification and regression tasks.

The overall basic process is as follows. The federated learning system is initiated by one server or participant, which sends an initial model to be trained by each participant with their own local data, who in turn delivers the weights or the gradients of the model back to the server (or to all the participants) to be aggregated. Then the federated model is sent back to the participants in an iterative way [1, 2]. The proceeding goes on until the termination conditions are accomplished. After this process, the output is a federated model that has been trained with the private data of all the participants [3].

This approach becomes critical when dealing with sensitive data, for instance in domains such as healthcare or finance. The aim of this paper, not an empirical research, is to propose an innovative federated learning approach for Fuzzy Cognitive Maps and to prove how appropriate FCMs are for Distributed Artificial Intelligence.

The proposal does not prioritise a particular optimisation method. In fact, this paper's primary emphasis is not on the training of FCMs, nor on the distributed training of FCMs.

Instead, the central focus of this paper is on the FCMs distributed training without an initial model. The main contributions of this paper are three-fold:

1. A privacy-preserving machine learning approach for FCMs. The authors design a training scheme for collaborative FCM training that includes data privacy. This proposal allows multiple participants to train an FCM model with their own data in compliance with strict data privacy regulations.
2. Two approaches to Fuzzy Cognitive Maps distributed learning. The authors propose two Particle Swarm Optimization-based FCM learning approach in a distributed way.
3. Blind Federated Learning as a new federated learning approach without an initial model, since the use of FCMs allows the participants not to define a model. To the best of our knowledge, this is the first federated learning proposal in which an initial model is not needed at all, defined neither by a server nor by the participants.

The authors test the validity of the proposal with well-known open datasets. The results of the experiments show that the proposal achieves a similar performance to the non-distributed method and improves the performance of the non-collaborative approach.

The rest of this paper is organized along these lines. We discuss the theoretical background in Section "Theoretical background". The methodological proposal is outlined in Section "Methodological proposal". Section "Experimental approach" describes the details of the experimental approach and the results. Finally, the authors draw the conclusions in Section "Conclusions".

## Theoretical background
### Fuzzy Cognitive Maps
*Fundamentals*

Fuzzy Cognitive Maps' nodes are modelling concepts, variables or features, the edges model relationships between them, and the weights represents the influence of those relations [4]. The value of a weight $\varpi_{ij}$ models how much node $c_i$ impacts over the node $c_j$. The fuzzy weights between edges are normalised within the ranges $\xi = \{[0, +1]|[-1, +1]\}$, depending if it includes just positive values or both positive and negative. The maximum positive influence is $+1.0$ and the opposite influence is $0.0$ or $-1.0$. The zero value shows that there is no correlation between the nodes. From a computational point of view, FCMs models are represented by a weight (adjacency) matrix which contains all edges' weights between the nodes.

The state of the nodes is shown as a state vector $c = [c_1, c_2, \ldots, c_N]$ that gives a snapshot of the states of the nodes at any iteration in the FCM dynamics [5]. The state of the node $i$ in the vector state at time $t$ (denoted as $c_i(t)$) is computed as shown in Eq. (1):

$$c_i(t) = f\left(\sum_{j=1}^{n} \varpi_{ji} \cdot c_j(t-1)\right) \qquad (1)$$

where $c_j$ are the presynaptic nodes and $\varpi_{ji}$ is the weight of the edge from $c_j$ to $c_i$. In a more formal way, a FCM can be denoted as a 4-tuple $\Phi = \langle c, \mathcal{W}, f, \xi \rangle$, where $c = \{c_i\}_{i=1}^{n}$ is the nodes' state with $n$ number of nodes, $\mathcal{W} = [\varpi_{ij}]_{n \times n} | -1, 0 \leq i, j \leq +1$ is the adjacency matrix representing the weights between the nodes, $f(\cdot)$ is the activation function, and $\xi$ is the nodes' range [6].

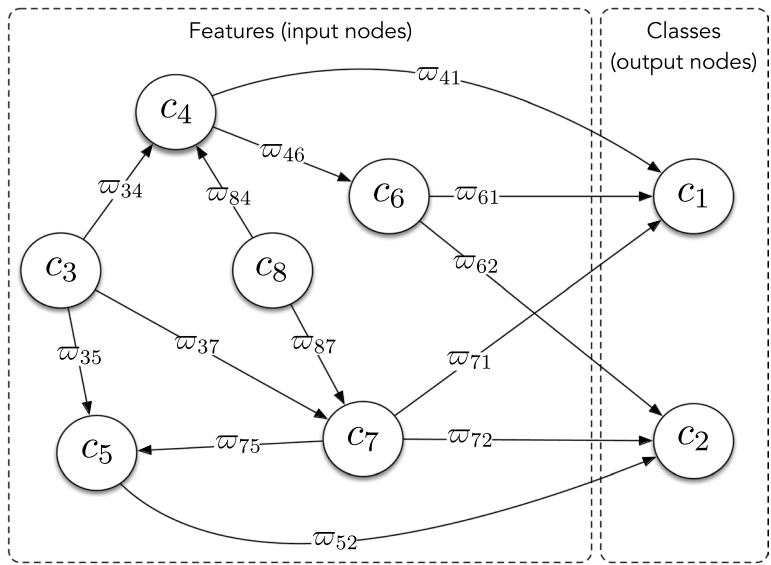

**Fig. 1** FCM binary classifier example

The FCM dynamical analysis begins with an initial vector state $c(0) = [c_1(0), \ldots, c_n(0)]$, which models the initial state of each node. The state of the nodes is updated in an iterative process. Thus, it includes a activation (transformation) function [7] for mapping monotonically the state of the node into a normalized range between $[0, +1]$ for unipolar FCMs or $[-1, +1]$ for bipolar ones. If the range is $[0, +1]$, the sigmoid is the most used transformation function, while hyperbolic tangent is the most used when the nodes' range is $[-1, +1]$ [8].

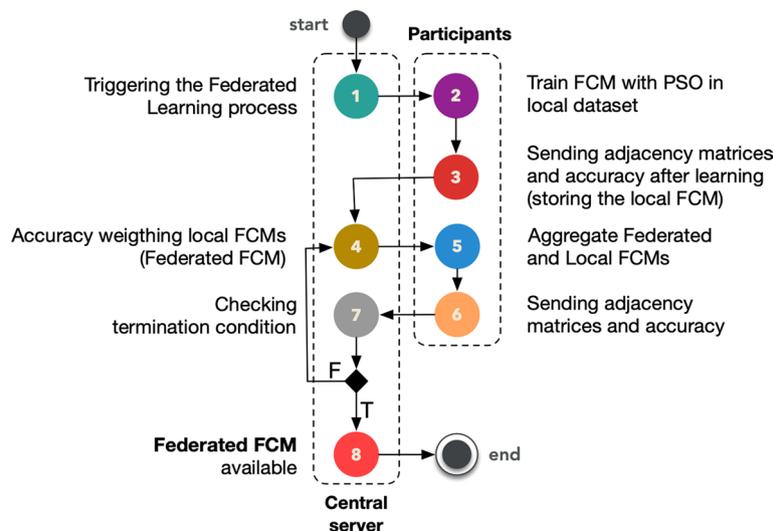

**Fig. 2** Proposed methodology

**Table 1** FCM no Federation, Breast Cancer dataset

| Agent | Size | % 1s | Accuracy | Precision |
|---|---|---|---|---|
| 1 | 100% | 37% | 0.9211 | 0.7742 |

**Table 2** FCM no Federation, Breast Cancer dataset, slope 5, sigmoid

| Agent | Size | % 1s | Accuracy | Precision |
|---|---|---|---|---|
| 1 | 100% | 37% | 0.8246 | 0.5714 |

If the selected activation function $f(\cdot)$ is the unipolar sigmoid, then the component $i$ of the vector state $c_i(t)$ at the instant $t$ is computed as shown in Eq. (2):

$$c_i(t) = \frac{1}{1 + e^{-\lambda \cdot \sum_{j=1}^{n} \varpi_{ji} \cdot c_j(t-1)}} \quad (2)$$

where $\lambda$ represents the slope of the unipolar sigmoid function. On the contrary, if the selected activation function $f(\cdot)$ is the hyperbolic tangent, then the node's state $c_i(t)$ at the instant $t$ is computed as Eq. (3) shows:

$$c_i(t) = \frac{\sinh\left(\lambda \cdot \sum_{j=1}^{n} \varpi_{ji} \cdot c_j(t-1)\right)}{\cosh\left(\lambda \cdot \sum_{j=1}^{n} \varpi_{ji} \cdot c_j(t-1)\right)} \quad (3)$$

After the dynamics, the FCM reaches one of three possible states after a number of iterations: it settles down to either a fixed pattern of node values (the so-called hidden pattern), to a limited cycle, or to a fixed-point attractor [9, 10].

### *Augmented FCMs*

There are two approaches to build FCMs. The first is through human experts [9]. This approach involves having each expert contribute their own FCM model. A group of experts should be carefully selected. Each expert individually design a FCM model that represents their own knowledge on the system to model. The second approach is automatic construction directly from raw data [5, 6, 11]. Due to the purpose of this research, this paper focuses on the distributed automatic construction of FCM.

According to the literature [7], an augmented adjacency matrix could be built by aggregating the adjacency matrix of each FCM. The elements' aggregation depends on whether there exist common nodes. If the adjacency matrices have common nodes, the states $\varpi_{jk}$ in the augmented matrix are computed by adding the adjacency matrix of each FCM model ($\mathcal{W}_i$).

The addition method when the adjacency matrices have not common nodes is known as direct sum of matrices, and the augmented matrix is denoted as $\odot_{i=1}^{N} \mathcal{W}_i$. Given a couple of FCMs with no common nodes and even different number of nodes with adjacency matrices $\varpi_{n \times n}^{A}$ and $\varpi_{m \times m}^{B}$, the resulting augmented adjacency matrix can be computed as in Eq. (4):

**Table 3** Blind FL results (constant weights)—Breast Cancer dataset

| Agent | Size | % 1s | Accuracy Pre-FL | Accuracy Post-FL | Precision Pre-FL | Precision Post-FL |
|---|---|---|---|---|---|---|
| 1 | 20% | 14% | 0.7727 | 0.9091 | 0.3750 | 0.6000 |
| 2 | 20% | 26% | 0.8696 | 0.9130 | 0.7143 | 0.8333 |
| 3 | 20% | 35% | 0.6957 | 0.8261 | 0.5385 | 0.7000 |
| 4 | 20% | 13% | 0.9130 | 0.9565 | 0.6667 | 0.5000 |
| 5 | 20% | 26% | 0.9565 | 0.9565 | 0.8571 | 0.8571 |
| Avg. | – | – | 0.8415 | **0.9123** | 0.6303 | **0.6981** |
| 1 | 19% | 24% | 0.9091 | 0.9318 | 0.8750 | 0.8889 |
| 2 | 26% | 35% | 0.8667 | 0.8833 | 0.8421 | 0.8889 |
| 3 | 16% | 42% | 0.8333 | 0.8611 | 1.0000 | 1.0000 |
| 4 | 15% | 20% | 0.8824 | 0.9118 | 0.6364 | 0.6364 |
| 5 | 24% | 47% | 0.8545 | 0.8727 | 0.9500 | 0.9524 |
| Avg. | – | – | 0.8692 | **0.8922** | 0.8607 | **0.8733** |
| 1 | 5% | 27% | 0.7500 | 0.7500 | 0.5000 | 0.5000 |
| 2 | 4% | 43% | 0.8000 | 0.8000 | 1.000 | 1.0000 |
| 3 | 42% | 35% | 0.9167 | 0.9271 | 0.9063 | 0.8824 |
| 4 | 20% | 23% | 0.9778 | 0.9778 | 0.9091 | 0.9091 |
| 5 | 29% | 49% | 0.7727 | 0.7727 | 0.9500 | 0.9048 |
| Avg. | – | – | 0.8434 | **0.8455** | 0.8531 | 0.8392 |
| 1 | 3% | 20% | 1.0000 | 1.0000 | 1.0000 | 0.5000 |
| 2 | 4% | 36% | 0.8000 | 0.8000 | 1.0000 | 1.0000 |
| 3 | 45% | 40% | 0.9327 | 0.9423 | 0.9268 | 0.9474 |
| 4 | 21% | 17% | 0.8542 | 0.8958 | 0.5454 | 0.6364 |
| 5 | 26% | 49% | 0.8361 | 0.8689 | 0.8889 | 0.9583 |
| Avg. | – | – | 0.8846 | **0.9014** | **0.8722** | 0.8084 |

Bold values indicate the best results

$$\bigodot_{i=1}^{N} \mathcal{W}_i = \begin{bmatrix} 0 & \mathcal{W}_{n \times n}^{A} \\ \mathcal{W}_{m \times m}^{B} & 0 \end{bmatrix} \quad (4)$$

where $N$ is the number of adjacency matrices to join, zeros are actually zero matrices, and the dimension of $\bigodot_{i=1}^{N} \mathcal{W}_i$ is $[\,\cdot\,]_{(m+n) \times (m+n)}$. In the case of common nodes, they would be computed as the average (or even the weighted average) of the nodes' states in each adjacency matrix $\mathcal{W}^i$.

*Pattern recognition with FCMs*

Because of the structure of an FCM model, it is a neuro-fuzzy technique and many concepts and procedures from neural networks can be applied in FCMs. FCMs have been applied both in classification and regression tasks. This paper is focused on the first task.

The literature has analysed pattern recognition tasks using Fuzzy Cognitive Maps. Papakostas et al. [12] and Papakostas and Koulouriotis [13] propose several FCM architectures for pattern recognition. Swzed [14] proposed a FCM based classifier with a fully connected architecture. Wu et al. [15] applied broad learning systems for time series classification with FCMs. Ramirez-Bautista et al. [16] applies FCMs for

**Table 4** Blind FL results (constant weights, slope 5, sigmoid)—Breast Cancer dataset

| Agent | Size | % 1s | Accuracy Pre-FL | Accuracy Post-FL | Precision Pre-FL | Precision Post-FL |
|---|---|---|---|---|---|---|
| 1 | 20% | 14% | 0.6818 | 0.9091 | 0.3000 | 0.6000 |
| 2 | 20% | 26% | 0.8696 | 0.8696 | 0.7143 | 0.6250 |
| 3 | 20% | 35% | 0.8696 | 0.8696 | 0.7778 | 0.7778 |
| 4 | 20% | 13% | 0.9130 | 0.9565 | 0.6667 | 0.5000 |
| 5 | 20% | 26% | 0.8696 | 0.9130 | 0.6667 | 0.7500 |
| Avg. | – | – | 0.8407 | **0.9036** | 0.6251 | **0.6506** |
| 1 | 19% | 24% | 0.9063 | 0.9063 | 0.6667 | 0.6667 |
| 2 | 26% | 35% | 0.9048 | 0.9365 | 0.9130 | 0.8846 |
| 3 | 16% | 42% | 0.7317 | 0.7317 | 1.0000 | 0.9167 |
| 4 | 15% | 20% | 0.9231 | 0.9231 | 0.7500 | 0.7500 |
| 5 | 24% | 47% | 0.8889 | 0.8889 | 0.9091 | 0.9091 |
| Avg. | – | – | 0.8709 | **0.8773** | **0.8478** | 0.8254 |
| 1 | 5% | 27% | 0.9167 | 0.9167 | 0.8333 | 0.8333 |
| 2 | 4% | 43% | 0.8667 | 0.8667 | 1.0000 | 1.0000 |
| 3 | 42% | 35% | 0.9300 | 0.9300 | 0.9189 | 0.8919 |
| 4 | 20% | 23% | 0.9487 | 0.9487 | 0.8571 | 0.8571 |
| 5 | 29% | 49% | 0.8571 | 0.8571 | 0.9474 | 0.9474 |
| Avg. | – | – | 0.9038 | 0.9038 | **0.9114** | 0.9059 |
| 1 | 3% | 20% | 0.8182 | 0.8182 | 0.3333 | 0.2500 |
| 2 | 4% | 36% | 0.7273 | 0.8182 | 0.5714 | 0.5714 |
| 3 | 45% | 40% | 0.8660 | 0.8763 | 0.8824 | 0.9032 |
| 4 | 21% | 17% | 0.9000 | 0.9250 | 0.7000 | 0.7778 |
| 5 | 26% | 49% | 0.9155 | 0.9155 | 1.0000 | 1.0000 |
| Avg. | – | – | 0.8454 | **0.8706** | 0.6974 | **0.7005** |

Bold values indicate the best results

classification of human plantar foot alterations. Baykasoglu and Golcuk [17] proposed alpha-cut based FCM methods are tested on several case studies. Papakostas et al. [18] applied unsupervised hebbian learning for pattern recognition problems.

In general terms, the main goal of a conventional classifier is the mapping of an input to a specific output according to a pattern. In this proposal, the input concepts represent the features of the dataset, while the output are the classes' labels where the patterns belong. Figure 1 shows an example topology of a Fuzzy Cognitive Map classifier, where the state of the concepts $c_1$ and $c_2$ defines the class where the input vector state belongs.

In that sense, if $c_1 > c_2$ the input vector state belongs to class 1, while if $c_1 < c_2$ the input vector state belongs to class 2. Note that if $c_1 = 0.03$ and $c_2 = 0.8$, then the input vector state would belong to class 2.

### FCM automatic construction with Particle Swarm Optimisation

FCM automatic construction endeavours are commonly focused on building the adjacency matrix based either on the available historical raw data or on expert knowledge [19–21]. FCM learning approaches could be divided into three categories [11, 22]: Hebbian, population-based, and hybrid mixing the main aspects of Hebbian-based and population-based learning algorithm.

**Table 5** Blended Blind FL results (constant weights) - Breast Cancer dataset

| Agent | Size | % 1s | Accuracy Pre-FL | Accuracy Post-FL | Precision Pre-FL | Precision Post-FL |
|---|---|---|---|---|---|---|
| 1 | 20% | 14% | 0.6818 | 0.9091 | 0.3000 | 0.6000 |
| 2 | 20% | 26% | 0.9130 | 0.9130 | 0.8333 | 0.8333 |
| 3 | 20% | 35% | 0.8696 | 0.8696 | 0.7778 | 0.7778 |
| 4 | 20% | 13% | 0.9130 | 0.9565 | 0.6667 | 0.5000 |
| 5 | 20% | 26% | 0.9565 | 1.0000 | 0.8571 | 1.0000 |
| Avg. | – | – | 0.8668 | **0.9296** | 0.6870 | **0.7422** |
| 1 | 19% | 24% | 0.9091 | 0.9318 | 0.8750 | 0.8889 |
| 2 | 26% | 35% | 0.8667 | 0.9000 | 0.8421 | 0.8889 |
| 3 | 16% | 42% | 0.8611 | 0.8611 | 1.0000 | 1.0000 |
| 4 | 15% | 20% | 0.8824 | 0.8824 | 0.6364 | 0.6364 |
| 5 | 24% | 47% | 0.8364 | 0.8728 | 0.9474 | 0.9524 |
| Avg. | – | – | 0.8711 | **0.8896** | 0.8602 | **0.8733** |
| 1 | 5% | 27% | 0.6667 | 0.7500 | 0.4000 | 0.5000 |
| 2 | 4% | 43% | 0.7000 | 0.7000 | 1.0000 | 1.0000 |
| 3 | 42% | 35% | 0.9479 | 0.9479 | 0.9394 | 0.8824 |
| 4 | 20% | 23% | 0.9778 | 0.9778 | 0.9091 | 0.9091 |
| 5 | 29% | 49% | 0.7576 | 0.7576 | 0.9048 | 0.9048 |
| Avg. | – | – | 0.8100 | **0.8267** | 0.8306 | **0.8392** |
| 1 | 3% | 20% | 0.8333 | 0.8333 | 0.5000 | 0.5000 |
| 2 | 4% | 36% | 0.8000 | 0.8000 | 1.0000 | 1.0000 |
| 3 | 45% | 40% | 0.9519 | 0.9519 | 0.9744 | 0.9500 |
| 4 | 21% | 17% | 0.8750 | 0.8958 | 0.5833 | 0.6364 |
| 5 | 26% | 49% | 0.8033 | 0.8689 | 0.9130 | 0.9583 |
| Avg. | – | – | 0.8527 | **0.8700** | 0.7941 | **0.8089** |

Bold values indicate the best results

The aim of the Hebbian-based FCM learning approaches is to modify adjacency matrices leading the FCM model to either achieve a steady state or converge into an acceptable region for the target system. This course has not been successful for FCMs extensions such as Fuzzy Grey Cognitive Maps [10].

Population-based methods do not need the human intervention. They compute adjacency matrices from historical raw data that best fit the sequence of input state vectors (the instances of the dataset). The learning goal of FCM evolutionary learning is to generate optimal adjacency matrix for modeling systems behaviour.

Particle Swarm Optimization is a bio-inspired, population-based and stochastic optimisation algorithm. The PSO algorithm generates a swarm of particles moving in an $n$-dimensional search space which must include all potential candidate solutions. In order to train the FCM adjacency matrices, we take into account the $k^{th}$ particle's position (a candidate solution or adjacency matrix), denoted as $\varpi_k = [\varpi_{k_1}, \ldots, \varpi_{k_j}]$ and its velocity, $v_k = [v_{k_1}, \ldots, v_{k_j}]$. Note that each particle is a potential solution or FCM candidate and its position $\varpi_k$ represents its adjacency matrix of the $k$-th FCM candidate [6, 23]. Each particle's velocity and position are updated at each time step. The position and the velocity of each particle are computed as shown in Eqs. (5) and (6):

$$\varpi_k(t+1) = \varpi_k(t) + v_k(t) \qquad (5)$$

**Table 6** Blended Blind FL results (constant weights, slope 5, sigmoid)—Breast Cancer dataset

| Agent | Size | % 1s | Accuracy Pre-FL | Accuracy Post-FL | Precision Pre-FL | Precision Post-FL |
|---|---|---|---|---|---|---|
| 1 | 20% | 14% | 0.6818 | 0.8636 | 0.3000 | 0.5000 |
| 2 | 20% | 26% | 0.8696 | 0.8696 | 0.7143 | 0.7143 |
| 3 | 20% | 35% | 0.7826 | 0.8261 | 0.6364 | 0.7000 |
| 4 | 20% | 13% | 0.8261 | 0.8261 | 0.4286 | 0.4286 |
| 5 | 20% | 26% | 0.9565 | 0.9565 | 0.8571 | 0.6667 |
| Avg. | – | – | 0.8233 | **0.8684** | 0.5873 | **0.6019** |
| 1 | 19% | 24% | 0.9063 | 0.9063 | 0.6667 | 0.6667 |
| 2 | 26% | 35% | 0.9365 | 0.9365 | 0.9200 | 0.8846 |
| 3 | 16% | 42% | 0.7073 | 0.7073 | 0.9231 | 0.9231 |
| 4 | 15% | 20% | 0.9487 | 0.9487 | 0.8571 | 0.7500 |
| 5 | 24% | 47% | 0.8704 | 0.8889 | 0.8696 | 0.9091 |
| Avg. | – | – | 0.8738 | **0.8775** | **0.8473** | 0.8267 |
| 1 | 5% | 27% | 1.0000 | 1.0000 | 1.0000 | 1.0000 |
| 2 | 4% | 43% | 0.9231 | 1.0000 | 0.7500 | 1.0000 |
| 3 | 42% | 35% | 0.9495 | 0.9495 | 0.9697 | 0.9697 |
| 4 | 20% | 23% | 0.9474 | 0.9474 | 0.7778 | 0.7778 |
| 5 | 29% | 49% | 0.8788 | 0.8788 | 1.0000 | 1.0000 |
| Avg. | – | – | 0.9397 | **0.9551** | 0.8995 | **0.9495** |
| 1 | 3% | 20% | 0.7778 | 0.8889 | 0.5000 | 0.5000 |
| 2 | 4% | 36% | 0.9091 | 0.9091 | 1.0000 | 1.0000 |
| 3 | 45% | 40% | 0.8785 | 0.8879 | 0.9375 | 0.9394 |
| 4 | 21% | 17% | 0.9524 | 0.9762 | 0.8333 | 1.0000 |
| 5 | 26% | 49% | 0.8833 | 0.9000 | 0.9310 | 0.9615 |
| Avg. | – | – | 0.8802 | **0.9124** | 0.8404 | **0.8802** |

Bold values indicate the best results

$$v_k(t+1) = v_k(t) + U(0,\phi_1) \otimes (\dot{\varpi}_k - \varpi_k(t)) + U(0,\phi_2) \otimes (\ddot{\varpi}_k - \varpi_k(t)) \qquad (6)$$

where $U(0, \phi_i)$ is a vector of random numbers from a uniform distribution within $[0, \phi_i]$, generated at each iteration and for each particle. Also, $\dot{\varpi}_k$ is the best position of particle $k$ in all former iterations, $\ddot{\varpi}_k$ is the best position of the whole population in all previous iterations, and $\otimes$ is the component-wise multiplication.

The PSO algorithm's goal is to locate all the particles in the global optima to a multidimensional hyper-volume. The fitness function used in this research is the complement of the Jaccard similarity coefficient ($\bar{J} = (Y \times \hat{Y}) \setminus J$). The Jaccard score computes the average of Jaccard similarity coefficients between pairs of the $i$-th samples, with a ground truth label set and a predicted label set. The complement operation is needed in terms of minimization of the fitness function. The Jaccard similarity coefficient's complement is computed as follows in Eq. (7):

$$\bar{J}(y_i, \hat{y}_i) = 1 - \frac{|y_i \cap \hat{y}_i|}{|y_i \cup \hat{y}_i|} \qquad (7)$$

The fitness function is sampled after each particle position update and is the objective function used to compute how close a given particle is in order to be able to achieve the global optimum.

**Table 7** Blind FL results (accuracy-based weights)—Breast Cancer dataset

| Agent | Size | % 1s | Accuracy Pre-FL | Accuracy Post-FL | Precision Pre-FL | Precision Post-FL |
|---|---|---|---|---|---|---|
| 1 | 20% | 14% | 0.7727 | 0.9091 | 0.3750 | 0.6000 |
| 2 | 20% | 26% | 0.9565 | 0.9565 | 0.8571 | 0.8333 |
| 3 | 20% | 35% | 0.8696 | 0.8696 | 0.7778 | 0.7000 |
| 4 | 20% | 13% | 0.8261 | 0.8696 | 0.4286 | 0.5000 |
| 5 | 20% | 26% | 1.0000 | 1.0000 | 1.0000 | 0.8571 |
| Avg. | – | – | 0.8859 | **0.9209** | 0.6877 | **0.6981** |
| 1 | 19% | 24% | 0.9318 | 0.9545 | 0.8182 | 0.8889 |
| 2 | 26% | 35% | 0.8500 | 0.8667 | 0.8750 | 0.8824 |
| 3 | 16% | 42% | 0.8611 | 0.8889 | 0.9444 | 1.0000 |
| 4 | 15% | 20% | 0.9118 | 0.9412 | 0.7500 | 0.5833 |
| 5 | 24% | 47% | 0.8909 | 0.8909 | 0.9545 | 0.9500 |
| Avg. | – | – | 0.8891 | **0.9084** | 0.8684 | 0.8609 |
| 1 | 5% | 27% | 0.7500 | 0.7500 | 0.5000 | 0.5000 |
| 2 | 4% | 43% | 0.7000 | 0.8000 | 1.0000 | 1.0000 |
| 3 | 42% | 35% | 0.9375 | 0.9375 | 0.9375 | 0.8824 |
| 4 | 20% | 23% | 0.9778 | 0.9778 | 0.9091 | 0.9091 |
| 5 | 29% | 49% | 0.7273 | 0.7576 | 0.8571 | 0.9048 |
| Avg. | – | – | 0.8185 | **0.8446** | 0.8407 | 0.8392 |
| 1 | 3% | 20% | 0.8333 | 0.8333 | 0.5000 | 0.5000 |
| 2 | 4% | 36% | 0.8000 | 0.9000 | 1.0000 | 1.0000 |
| 3 | 45% | 40% | 0.9423 | 0.9423 | 0.9500 | 0.9487 |
| 4 | 21% | 17% | 0.9167 | 0.9167 | 0.7000 | 0.6364 |
| 5 | 26% | 49% | 0.8689 | 0.8689 | 1.0000 | 1.0000 |
| Avg. | – | – | 0.8722 | **0.8922** | **0.8300** | 0.8170 |

Bold values indicate the best results

### Federated learning

Distributed Artificial Intelligence is the subfield of artificial intelligence that studies the sharing of knowledge between agents in order to solve complex problems, classically via the distribution of tasks or data. Such processes may not be of interest in fields where the characteristics of the data and the regulations make it impossible to share it, such as finance or health.

Conventional machine learning requires all data collected on local devices to be stored centrally on a data silo. The goal of federated learning is building a global model that can be trained on data distributed while assuring the data privacy [24]. Federated learning is one of the most recent efforts in secure distributed artificial intelligence, proposed by McMahan et al. [2] and further developed in Konecny et al. [25] and McMahan and Ramage [26]. Some advantages of federated learning are privacy protection and the possibility of solving complex problems with small data samples such as healthcare [27].

In the recent years, there have been several attempts to create a federated version of conventional machine learning algorithms, such as federated linear regression [28–30], federated logistic regression [31], federated random forest [32], federated XGBoost [33–35], and federated support vector machines [36, 37]. To the best of our knowledge, this is the first work focusing on utilizing FCM in a blind federated setting.

A centralised federated learning system can be described as follows:

**Table 8** Blind FL results (accuracy-based weights, slope 5, sigmoid)—Breast Cancer dataset

| Agent | Size | % 1s | Accuracy Pre-FL | Accuracy Post-FL | Precision Pre-FL | Precision Post-FL |
|---|---|---|---|---|---|---|
| 1 | 20% | 14% | 0.8636 | 0.8636 | 0.5000 | 0.5000 |
| 2 | 20% | 26% | 0.9130 | 0.9139 | 0.8333 | 0.7143 |
| 3 | 20% | 35% | 0.8696 | 0.8696 | 0.7778 | 0.7000 |
| 4 | 20% | 13% | 0.7826 | 0.8696 | 0.2500 | 0.4286 |
| 5 | 20% | 26% | 0.8261 | 0.9130 | 0.6000 | 0.7500 |
| Avg. | – | – | 0.8510 | **0.8858** | 0.5922 | **0.6186** |
| 1 | 19% | 24% | 0.9063 | 0.9063 | 0.6667 | 0.6667 |
| 2 | 26% | 35% | 0.9048 | 0.9048 | 0.8519 | 0.8519 |
| 3 | 16% | 42% | 0.7317 | 0.7317 | 1.0000 | 1.0000 |
| 4 | 15% | 20% | 0.9231 | 0.9231 | 0.7500 | 0.7500 |
| 5 | 24% | 47% | 0.8519 | 0.8704 | 0.8636 | 0.8696 |
| Avg. | – | – | 0.8635 | **0.8672** | 0.8264 | **0.8276** |
| 1 | 5% | 27% | 1.0000 | 1.0000 | 1.0000 | 1.0000 |
| 2 | 4% | 43% | 0.8182 | 0.8182 | 1.0000 | 1.0000 |
| 3 | 42% | 35% | 0.9293 | 0.9394 | 0.9706 | 0.9375 |
| 4 | 20% | 23% | 0.8889 | 0.8889 | 0.7000 | 0.7000 |
| 5 | 29% | 49% | 0.9028 | 0.9028 | 0.9600 | 0.9583 |
| Avg. | – | – | 0.9078 | **0.9098** | **0.9261** | 0.9192 |
| 1 | 3% | 20% | 1.0000 | 1.0000 | 1.0000 | 1.0000 |
| 2 | 4% | 36% | 0.9231 | 0.9231 | 1.0000 | 1.0000 |
| 3 | 45% | 40% | 0.9515 | 0.9515 | 0.9737 | 0.9459 |
| 4 | 21% | 17% | 0.9512 | 0.9512 | 0.8889 | 0.8889 |
| 5 | 26% | 49% | 0.9219 | 0.9375 | 0.9615 | 1.0000 |
| Avg. | – | – | 0.9495 | **0.9527** | 0.9648 | **0.9670** |

Bold values indicate the best results

1. The central server delivers a model to each agent. In the initial iteration of this process, the server has built an empty model.
2. The participants train the model with their own private data.
3. Each participant sends the parameters of the model or its gradients to the central server in a private way, usually encrypted.
4. The central server builds a federated model by aggregating the parameters of the individual models.
5. The central server checks if the termination condition is accomplished in which case the federated model is finished, otherwise the process goes back to step 1.

The ultimate goal of the federated model is to minimize the total loss (Eq. 8) of all participants computed as follows:

$$\mathcal{L}^* = \sum_{i=1}^{n} \kappa_i \cdot \mathcal{L}(\mathcal{D}_i, \Psi) \qquad (8)$$

where $\Psi$ are the model parameters, $\mathcal{D}_i$ is the dataset of the participant $i$, $\mathcal{L}^*$ is the loss function of the federated model, $\mathcal{L}_i(\cdot)$ is the loss function for each participant in the federation, and $\kappa_i$ represent the importance (weight) of each participant. It is possible to determine $\kappa_i$ by several criteria such as dataset size, accuracy and so on.

**Table 9** Blended Blind FL results (accuracy-based weights)—Breast Cancer dataset

| Agent | Size | % 1s | Accuracy Pre-FL | Accuracy Post-FL | Precision Pre-FL | Precision Post-FL |
|---|---|---|---|---|---|---|
| 1 | 20% | 14% | 0.7273 | 0.9091 | 0.3333 | 0.6000 |
| 2 | 20% | 26% | 0.9130 | 0.9130 | 0.8333 | 0.8333 |
| 3 | 20% | 35% | 0.8696 | 0.8696 | 0.7778 | 0.7778 |
| 4 | 20% | 13% | 0.9130 | 0.9130 | 0.6000 | 0.5000 |
| 5 | 20% | 26% | 0.8696 | 1.0000 | 0.6666 | 1.0000 |
| Avg | – | – | 0.8585 | **0.9209** | 0.6422 | **0.7422** |
| 1 | 5% | 27% | 0.9545 | 0.9773 | 0.9000 | 1.0000 |
| 2 | 4% | 43% | 0.9167 | 0.9167 | 0.9000 | 0.8947 |
| 3 | 42% | 35% | 0.8611 | 0.8611 | 1.0000 | 1.0000 |
| 4 | 20% | 23% | 0.9118 | 0.9118 | 0.7000 | 0.5833 |
| 5 | 29% | 49% | 0.8909 | 0.8909 | 0.9545 | 0.9545 |
| Avg. | – | – | 0.9070 | **0.9115** | **0.8909** | 0.8865 |
| 1 | 19% | 24% | 0.6667 | 0.7500 | 0.4000 | 0.5000 |
| 2 | 26% | 35% | 0.8000 | 0.8000 | 1.0000 | 1.0000 |
| 3 | 16% | 42% | 0.9063 | 0.9167 | 0.8571 | 0.8824 |
| 4 | 15% | 20% | 0.9778 | 0.9778 | 0.9091 | 0.9091 |
| 5 | 24% | 47% | 0.7576 | 0.7576 | 0.9048 | 0.9048 |
| Avg. | – | – | 0.8217 | **0.8404** | 0.8142 | **0.8392** |
| 1 | 3% | 20% | 0.8333 | 1.0000 | 0.5000 | 1.0000 |
| 2 | 4% | 36% | 0.9000 | 0.9000 | 1.0000 | 1.0000 |
| 3 | 45% | 40% | 0.9615 | 0.9615 | 0.9750 | 0.9487 |
| 4 | 21% | 17% | 0.8333 | 0.8958 | 0.5000 | 0.6364 |
| 5 | 26% | 49% | 0.8525 | 0.8525 | 1.0000 | 1.0000 |
| Avg. | – | – | 0.8761 | **0.9220** | 0.7950 | **0.9170** |

Bold values indicate the best results

The first application of federated learning was to create collaborative predictive models using private data in Android mobile phones [26]. In particular, a model in Gboard on Android, the Google Keyboard, in order to predict the following word or phrase that the user is going to write based on the former text and other users (private) data. In this set-up the central server manages the federated model and the communications with the agents, while the participants own their data and train the partial models. In this way, a federated learning system ensures that the distributed model is built in a private environment, since the private data never leaves the local agent.

Nevertheless, there are always risks associated with the data transmission, such as the possibility of the reconstruction of the model or the training data from the model parameters. Due to these risks, there is an increasing interest in the use of an additional layer of privacy to this information, and there are many studies that use privacy-preserving methods in federated learning such as Differential Privacy [38], Secure Multi-Party Computation [39] or Homomorphic encryption [40]. The comparison with other privacy-preserving techniques is outside the scope of this work, focused on the construction of a federation process using FCMs and without an initial model, but in the philosophy of federated learning, an extra security layer, such as Differential Privacy, could be added at the time of sharing the parameters of the model [41].

**Table 10** Blended Blind FL results (accuracy-based weights, slope 5, sigmoid)—Breast Cancer dataset

| Agent | Size | % 1s | Accuracy Pre-FL | Accuracy Post-FL | Precision Pre-FL | Precision Post-FL |
|---|---|---|---|---|---|---|
| 1 | 20% | 14% | 0.7727 | 0.9091 | 0.3750 | 0.6000 |
| 2 | 20% | 26% | 0.9130 | 0.9130 | 0.8333 | 0.8333 |
| 3 | 20% | 35% | 0.7391 | 0.9130 | 0.5714 | 0.7778 |
| 4 | 20% | 13% | 0.9565 | 0.9565 | 0.7500 | 0.5000 |
| 5 | 20% | 26% | 0.9565 | 1.0000 | 1.0000 | 1.0000 |
| Avg. | – | – | 0.8676 | **0.9383** | 0.7060 | **0.7422** |
| 1 | 5% | 27% | 0.9063 | 0.9063 | 0.6667 | 0.6667 |
| 2 | 4% | 43% | 0.9524 | 0.9524 | 0.9583 | 0.8846 |
| 3 | 42% | 35% | 0.7317 | 0.7317 | 1.0000 | 1.0000 |
| 4 | 20% | 23% | 0.9231 | 0.9231 | 0.7500 | 0.7500 |
| 5 | 29% | 49% | 0.8519 | 0.8704 | 0.8636 | 0.9048 |
| Avg. | – | – | 0.8731 | **0.8768** | **0.8477** | 0.8412 |
| 1 | 19% | 24% | 0.9167 | 0.9167 | 1.0000 | 0.8000 |
| 2 | 26% | 35% | 0.9286 | 0.9286 | 1.0000 | 1.0000 |
| 3 | 16% | 42% | 0.8990 | 0.9091 | 0.9643 | 0.8750 |
| 4 | 15% | 20% | 0.9268 | 0.9268 | 0.8750 | 0.8750 |
| 5 | 24% | 47% | 0.8730 | 0.8889 | 0.9565 | 1.0000 |
| Avg. | – | – | 0.9088 | **0.9140** | **0.9592** | 0.9100 |
| 1 | 3% | 20% | 1.0000 | 1.0000 | 1.0000 | 1.0000 |
| 2 | 4% | 36% | 0.9231 | 0.9231 | 1.0000 | 1.0000 |
| 3 | 45% | 40% | 0.8835 | 0.8932 | 0.9375 | 0.9394 |
| 4 | 21% | 17% | 0.9767 | 0.9767 | 0.9167 | 0.8462 |
| 5 | 26% | 49% | 0.9167 | 0.9333 | 0.9600 | 1.0000 |
| Avg. | – | – | 0.9400 | **0.9453** | **0.9628** | 0.9571 |

Bold values indicate the best results

Federated learning represents a significant step forward in the privacy-preserving machine learning field. Its practical managerial significance lies in its potential to address the balance between utilizing valuable data for business insights, respecting privacy regulations and customer trust. By allowing model distributed training on decentralised data sources while preserving privacy, federated learning offers several managerial benefits:

- Collaborative business insights: FL can facilitate collaboration between different business units or partners without sharing sensitive data directly. This fosters knowledge sharing and cross-functional collaboration while maintaining data privacy.
- Enhanced data privacy compliance: FL enables organisations to comply with strict data protection regulations such as GDPR. This approach avoids reputational damage that may result from non-compliance of data leaks .
- Cost-Efficient AI training: Since data remains on local devices or servers, it reduces the need for extensive data transfer and centralised storage infrastructure.
- Customer trust and brand loyalty: Companies can build trust with their customers by demonstrating a strong commitment to data privacy. This trust can lead to increased customer loyalty and positive brand perception.

Table 11 Blind FL results (Precision-based weights)—Breast Cancer dataset

| Agent | Size | % 1s | Accuracy Pre-FL | Accuracy Post-FL | Precision Pre-FL | Precision Post-FL |
|---|---|---|---|---|---|---|
| 1 | 20% | 14% | 0.7273 | 0.9091 | 0.3333 | 0.6000 |
| 2 | 20% | 26% | 0.8261 | 0.9130 | 0.6250 | 0.8333 |
| 3 | 20% | 35% | 0.7826 | 0.8696 | 0.6364 | 0.7778 |
| 4 | 20% | 13% | 0.8696 | 0.9130 | 0.5000 | 0.5000 |
| 5 | 20% | 26% | 0.9696 | 0.9565 | 0.6667 | 0.8571 |
| Avg. | – | – | 0.8150 | **0.9123** | 0.5523 | **0.7137** |
| 1 | 19% | 24% | 0.9318 | 0.9545 | 1.0000 | 0.8889 |
| 2 | 26% | 35% | 0.8667 | 0.8833 | 0.8421 | 0.8889 |
| 3 | 16% | 42% | 0.8889 | 0.8889 | 1.0000 | 1.0000 |
| 4 | 15% | 20% | 0.8824 | 0.8824 | 0.6364 | 0.6364 |
| 5 | 24% | 47% | 0.8727 | 0.8727 | 0.9524 | 0.9524 |
| Avg. | – | – | 0.8885 | **0.8964** | 0.8862 | **0.8733** |
| 1 | 5% | 27% | 0.6667 | 0.7500 | 0.4000 | 0.5000 |
| 2 | 4% | 43% | 0.8000 | 0.8000 | 1.0000 | 1.0000 |
| 3 | 42% | 35% | 0.9583 | 0.9688 | 1.0000 | 0.9091 |
| 4 | 20% | 23% | 0.9778 | 0.9778 | 0.9091 | 0.9091 |
| 5 | 29% | 49% | 0.7576 | 0.7879 | 0.9474 | 1.0000 |
| Avg. | – | – | 0.8321 | **0.8569** | 0.8513 | **0.8636** |
| 1 | 3% | 20% | 0.8333 | 0.8333 | 0.5000 | 0.5000 |
| 2 | 4% | 36% | 0.7000 | 0.8000 | 1.0000 | 1.0000 |
| 3 | 45% | 40% | 0.9423 | 0.9423 | 0.9500 | 0.9487 |
| 4 | 21% | 17% | 0.8958 | 0.9167 | 0.6364 | 0.5833 |
| 5 | 26% | 49% | 0.8689 | 0.8689 | 0.9600 | 0.9565 |
| Avg. | – | – | 0.8481 | **0.8722** | 0.8093 | 0.7977 |

Bold values indicate the best results

In this sense, a practical real-world healthcare FL application would involve a consortium of healthcare institutions or health data owners working together to improve patient care and disease prediction while preserving data privacy. In this scenario, each institution would retain control of its patient data, ensuring compliance with strict privacy regulations like HIPAA and GDPR.

In Section "Federated learning for FCMs", the authors detail the federated learning approach and the proposed methodology for FCMs, that enables the creation of a machine learning model between several agents while all the participants keep their data private.

## Methodological proposal

### Federated learning for FCMs

The proposed Blind Federated Learning methodology for FCMs is shown in Fig. 2, and can be described as follows:

1. Although the central server has no data, it triggers the Blind Federated Learning process by setting in motion the participants, who own the data to train the final model. Note that the central server does not send any initial FCM to the participants and this is one contribution of this research. As far as we know this is the first federated

**Table 12** Blind FL results (Precision-based weights, slope 5, sigmoid)—Breast Cancer dataset

| Agent | Size | % 1s | Accuracy Pre-FL | Accuracy Post-FL | Precision Pre-FL | Precision Post-FL |
|---|---|---|---|---|---|---|
| 1 | 20% | 14% | 0.5909 | 0.9091 | 0.2500 | 0.6000 |
| 2 | 20% | 26% | 0.9130 | 0.9130 | 0.8333 | 0.8333 |
| 3 | 20% | 35% | 0.8696 | 0.8696 | 0.7778 | 0.7778 |
| 4 | 20% | 13% | 0.7826 | 0.8696 | 0.3750 | 0.5000 |
| 5 | 20% | 26% | 0.9565 | 1.0000 | 0.8571 | 1.0000 |
| Avg. | – | – | 0.8225 | **0.9123** | 0.6187 | **0.7422** |
| 1 | 19% | 24% | 0.8750 | 0.8750 | 0.6000 | 0.6000 |
| 2 | 26% | 35% | 0.9048 | 0.9206 | 0.8800 | 0.8846 |
| 3 | 16% | 42% | 0.7317 | 0.7317 | 1.0000 | 1.0000 |
| 4 | 15% | 20% | 0.9231 | 0.9231 | 0.7500 | 0.7500 |
| 5 | 24% | 47% | 0.8519 | 0.8889 | 0.8636 | 0.9091 |
| Avg. | – | – | 0.8573 | **0.8679** | 0.8187 | **0.8287** |
| 1 | 5% | 27% | 0.9000 | 0.9000 | 0.7500 | 0.7500 |
| 2 | 4% | 43% | 0.9286 | 0.9286 | 0.8333 | 0.8333 |
| 3 | 42% | 35% | 0.9035 | 0.9035 | 0.9024 | 0.8810 |
| 4 | 20% | 23% | 0.8974 | 0.9231 | 0.7500 | 0.7500 |
| 5 | 29% | 49% | 0.9423 | 0.9423 | 1.0000 | 1.0000 |
| Avg. | – | – | 0.9144 | **0.9195** | **0.8472** | 0.8429 |
| 1 | 3% | 20% | 0.8333 | 0.8333 | 0.3333 | 0.3333 |
| 2 | 4% | 36% | 0.8571 | 0.9286 | 0.8000 | 0.8333 |
| 3 | 45% | 40% | 0.9107 | 0.9107 | 1.0000 | 0.9697 |
| 4 | 21% | 17% | 0.8857 | 0.8857 | 0.7143 | 0.7143 |
| 5 | 26% | 49% | 0.8621 | 0.8793 | 0.9583 | 0.9600 |
| Avg. | – | – | 0.8698 | **0.8875** | 0.7612 | **0.7621** |

Bold values indicate the best results

learning proposal that it does not need an initial model, then the central server is not even required.

2. Each participant trains its own initial local FCM with their own dataset. The authors have used a PSO algorithm to train the FCMs, and that the dynamics have converged when the difference between two consecutive vector states is under a tolerance value (in these experiments was $1 \times 10^{-5}$), but this proposal is agnostic to the learning approach and to other considerations.

3. Each participant delivers its model parameters, which in this case are the trained adjacency matrices. If needed, the participants could send any other performance metric needed to calculate the averaging of the models (see section "Aggregation methods"). Due to the privacy concerns discussed earlier, the parameters may be encrypted using a privacy-preserving method. Finally, the local FCM is stored in the participant devices.

4. The central server aggregates the parameters of the local models in its device using the appropriate weight. The section "Aggregation methods" shows a detailed description of the different aggregation methods considered by the authors. This process results in the parameters of a federated model.

5. The participants receive these parameters from the central server. They build the next iteration of their local model using the federated model parameters. The authors

**Table 13** Blended Blind FL results (Precision-based weights) - Breast Cancer dataset

| Agent | Size | % 1s | Accuracy Pre-FL | Accuracy Post-FL | Precision Pre-FL | Precision Post-FL |
|---|---|---|---|---|---|---|
| 1 | 20% | 14% | 0.8182 | 0.9091 | 0.4286 | 0.6000 |
| 2 | 20% | 26% | 0.9130 | 0.9130 | 0.8333 | 0.8333 |
| 3 | 20% | 35% | 0.8261 | 0.8696 | 0.7000 | 0.7778 |
| 4 | 20% | 13% | 0.9130 | 0.9130 | 0.6000 | 0.6000 |
| 5 | 20% | 26% | 0.9130 | 1.0000 | 0.7500 | 1.0000 |
| Avg. | – | – | 0.8767 | **0.9209** | 0.6624 | **0.7622** |
| 1 | 19% | 24% | 0.9545 | 0.9773 | 0.9000 | 0.8889 |
| 2 | 26% | 35% | 0.9000 | 0.9000 | 0.8947 | 0.8889 |
| 3 | 16% | 42% | 0.8333 | 0.8611 | 1.0000 | 1.0000 |
| 4 | 15% | 20% | 0.8529 | 0.8529 | 0.5833 | 0.5833 |
| 5 | 24% | 47% | 0.8727 | 0.8727 | 0.9524 | 0.9500 |
| Avg. | – | – | 0.8827 | **0.8928** | 0.8661 | 0.8622 |
| 1 | 5% | 27% | 0.7500 | 0.7500 | 0.5000 | 0.5000 |
| 2 | 4% | 43% | 0.8000 | 0.8000 | 1.0000 | 1.0000 |
| 3 | 42% | 35% | 0.9063 | 0.9167 | 0.8788 | 0.9063 |
| 4 | 20% | 23% | 0.9778 | 0.9778 | 0.9091 | 0.9091 |
| 5 | 29% | 49% | 0.7727 | 0.7727 | 0.9500 | 0.8261 |
| Avg. | – | – | 0.8414 | **0.8434** | 0.8476 | 0.8283 |
| 1 | 3% | 20% | 0.8333 | 0.8333 | 0.5000 | 0.5000 |
| 2 | 4% | 36% | 0.9000 | 0.9000 | 1.0000 | 1.0000 |
| 3 | 45% | 40% | 0.9519 | 0.9519 | 0.9512 | 0.9487 |
| 4 | 21% | 17% | 0.8333 | 0.8750 | 0.5000 | 0.5833 |
| 5 | 26% | 49% | 0.8361 | 0.8689 | 0.9565 | 1.0000 |
| Avg. | – | – | 0.8709 | **0.8858** | 0.7815 | **0.8064** |

Bold values indicate the best results

  propose two different federation methods: (a) in the proposal that is closer to the Blind Federated Learning, the local model is just the global model, which in turn was created as aggregation of all participant's models; and (b) an innovative federated learning approach, called Blended Blind Federated Learning, where the new local model is obtained by combining the new received adjacency matrix with the previous local adjacency matrix.

6. In either aggregation case, the participant retrains the new FCM in their local data and sends the parameters (and the needed performance metrics) back to the central server to be aggregated once again. Also, at this point the participants use their local data to test the performance of their local model.
7. The central server checks whether the termination condition is met. The authors have chosen that the federation process must be run 20 iterations. If the condition is not fulfilled, the process goes back to stage 4.
8. If the termination condition is satisfied, then the federated learning process ends resulting in a Federated FCM.

The proposed approach deals with the issue of federated learning without the need for an initial model. To the best of the authors' knowledge, this problem remains unsolved. For this reason, we view this paper as a valuable undertaking.

Table 14 Blended Blind FL results (Precision-based weights, slope 5, sigmoid)—Breast Cancer dataset

| Agent | Size | % 1s | Accuracy Pre-FL | Accuracy Post-FL | Precision Pre-FL | Precision Post-FL |
|---|---|---|---|---|---|---|
| 1 | 20% | 14% | 0.7727 | 0.9091 | 0.3750 | 0.6000 |
| 2 | 20% | 26% | 0.8696 | 0.9130 | 0.7143 | 0.8333 |
| 3 | 20% | 35% | 0.9565 | 0.9565 | 1.0000 | 0.7778 |
| 4 | 20% | 13% | 0.9130 | 0.9130 | 0.6000 | 0.5000 |
| 5 | 20% | 26% | 0.8696 | 0.9565 | 0.6667 | 0.8571 |
| Avg. | – | – | 0.8763 | **0.9296** | 0.6712 | **0.7137** |
| 1 | 19% | 24% | 0.8750 | 0.9063 | 0.6250 | 0.6667 |
| 2 | 26% | 35% | 0.9524 | 0.9524 | 0.9583 | 0.8519 |
| 3 | 16% | 42% | 0.7073 | 0.7317 | 0.9231 | 1.0000 |
| 4 | 15% | 20% | 0.9231 | 0.9231 | 0.7500 | 0.7500 |
| 5 | 24% | 47% | 0.8704 | 0.8889 | 0.8696 | 0.9091 |
| Avg. | – | – | 0.8656 | **0.8805** | 0.8252 | **0.8355** |
| 1 | 5% | 27% | 0.8462 | 0.8462 | 0.6667 | 0.6667 |
| 2 | 4% | 43% | 1.0000 | 1.0000 | 1.0000 | 1.0000 |
| 3 | 42% | 35% | 0.8969 | 0.8969 | 0.8684 | 0.8684 |
| 4 | 20% | 23% | 0.8974 | 0.9231 | 0.7500 | 0.8182 |
| 5 | 29% | 49% | 0.8986 | 0.8986 | 1.0000 | 1.0000 |
| Avg. | – | – | 0.9078 | **0.9129** | 0.8570 | **0.8707** |
| 1 | 3% | 20% | 0.8750 | 0.8750 | 0.6667 | 0.4000 |
| 2 | 4% | 36% | 0.8462 | 0.8462 | 1.0000 | 1.0000 |
| 3 | 45% | 40% | 0.8889 | 0.8990 | 0.8421 | 0.8421 |
| 4 | 21% | 17% | 0.9024 | 0.9024 | 0.5000 | 0.3636 |
| 5 | 26% | 49% | 0.8971 | 0.9118 | 1.0000 | 1.0000 |
| Avg. | – | – | 0.8819 | **0.8869** | 0.8018 | 0.7211 |

Bold values indicate the best results

Table 15 FCM no Federation—adult dataset

| Agent | Size | % 1s | Accuracy | Precision |
|---|---|---|---|---|
| 1 | 100% | 37% | 0.8611 | 0.9487 |

Table 16 FCM no Federation, Adult dataset, slope 5, sigmoid

| Agent | Size | % 1s | Accuracy | Precision |
|---|---|---|---|---|
| 1 | 100% | 37% | 0.9222 | 0.9257 |

**Aggregation methods**

An important parameter when defining a federated learning approach is the aggregation method employed by the central server to obtain the federated model.

In this paper, the authors propose three different approaches:

1. Federated averaging. This method performs the aggregation using the arithmetic average [2]. The central server sums the parameters of the different models and

**Table 17** Blind FL results (constant weights)—adult dataset

| Agent | Size | % 1s | Accuracy Pre-FL | Accuracy Post-FL | Precision Pre-FL | Precision Post-FL |
|---|---|---|---|---|---|---|
| 1 | 20% | 72% | 0.9444 | 0.9444 | 0.9615 | 0.9259 |
| 2 | 20% | 75% | 0.8055 | 0.8611 | 0.8846 | 0.9230 |
| 3 | 20% | 75% | 0.8611 | 0.9166 | 0.8928 | 0.9600 |
| 4 | 20% | 67% | 0.8888 | 0.9166 | 0.9545 | 0.9565 |
| 5 | 20% | 72% | 0.9444 | 0.9722 | 0.9615 | 1.0000 |
| Avg. | – | – | 0.8889 | **0.9222** | 0.9310 | **0.9531** |
| 1 | 13% | 63% | 0.5000 | 0.5000 | 1.0000 | 1.0000 |
| 2 | 30% | 80% | 0.3261 | 0.3478 | 1.0000 | 1.0000 |
| 3 | 21% | 87% | 0.2821 | 0.3077 | 1.0000 | 1.0000 |
| 4 | 11% | 61% | 0.6047 | 0.6279 | 1.0000 | 1.0000 |
| 5 | 24% | 81% | 0.3824 | 0.3922 | 0.9615 | 0.9412 |
| Avg. | – | – | 0.4190 | **0.4351** | **0.9923** | 0.9882 |
| 1 | 4% | 75% | 0.3125 | 0.3125 | 1.0000 | 0.0000 |
| 2 | 5% | 68% | 0.3684 | 0.3684 | 0.6666 | 0.6666 |
| 3 | 45% | 77% | 0.3496 | 0.3619 | 0.8571 | 0.8260 |
| 4 | 13% | 51% | 0.5869 | 0.5869 | 1.0000 | 1.0000 |
| 5 | 32% | 85% | 0.2649 | 0.2820 | 0.8823 | 0.8125 |
| Avg. | – | – | 0.3765 | **0.3823** | **0.8812** | 0.6610 |
| 1 | 1% | 25% | 0.3333 | 0.3333 | 0.0000 | 0.0000 |
| 2 | 5% | 72% | 0.3500 | 0.3500 | 1.0000 | 0.0000 |
| 3 | 45% | 78% | 0.2546 | 0.2732 | 0.8750 | 0.8750 |
| 4 | 14% | 53% | 0.4693 | 0.4693 | 0.0000 | 0.0000 |
| 5 | 35% | 89% | 0.1825 | 0.1904 | 1.0000 | 1.0000 |
| Avg. | – | – | 0.3179 | **0.3232** | **0.5750** | 0.3750 |

Bold values indicate the best results

divides by the number of participants (or models). This way, the federated model weighs all the participants in a similar fashion. The parameters of the federated model $\varpi^*$ would be computed as shown in Eq. (9):

$$\varpi^* = \frac{1}{n} \cdot \sum_{i=1}^{n} \varpi_i \qquad (9)$$

where $n$ is the number of participants and $\varpi_i$ are the parameters of the local model for participant $i$.

2. Accuracy-based federated weighted averaging, with the normalized accuracy of each model as the weight: The central server receives not only the individual models, but the accuracy of each model in a test set for the participant as well. Then, it averages the models parameters using a weighted average with the normalized accuracy for each participant as its weight. Therefore, the aggregation weighs reinforces the participant that contributes the most to the general accuracy. In this case, the parameters of the federated model $\varpi^*$ would be computed as follows in Eq. (10):

$$\varpi^* = \sum_{i=1}^{n} \psi_i \cdot \varpi_i, \qquad (10)$$

Table 18 Blind FL results (constant weights, slope 5, sigmoid)—adult dataset

| Agent | Size | % 1s | Accuracy Pre-FL | Accuracy Post-FL | Precision Pre-FL | Precision Post-FL |
|---|---|---|---|---|---|---|
| 1 | 20% | 72% | 0.9444 | 0.9722 | 0.9310 | 0.9310 |
| 2 | 20% | 75% | 0.8333 | 0.9722 | 0.8286 | 0.9667 |
| 3 | 20% | 75% | 0.8889 | 0.9167 | 0.8667 | 0.8966 |
| 4 | 20% | 67% | 0.9444 | 0.9722 | 0.9655 | 0.8788 |
| 5 | 20% | 72% | 0.9444 | 0.9722 | 1.0000 | 0.9333 |
| Avg. | – | – | 0.9111 | **0.9611** | 0.9184 | **0.9213** |
| 1 | 13% | 63% | 0.3227 | 0.3642 | 0.8571 | 0.9063 |
| 2 | 30% | 80% | 0.3008 | 0.3384 | 0.7500 | 1.0000 |
| 3 | 21% | 87% | 0.3324 | 0.3526 | 0.9048 | 0.9355 |
| 4 | 11% | 61% | 0.3701 | 0.4156 | 0.9091 | 0.9444 |
| 5 | 24% | 81% | 0.3479 | 0.3673 | 0.9398 | 0.9091 |
| Avg. | – | – | 0.3348 | **0.3676** | 0.8722 | **0.9391** |
| 1 | 4% | 75% | 0.2561 | 0.3049 | 0.7500 | 0.8333 |
| 2 | 5% | 68% | 0.4030 | 0.4328 | 1.0000 | 1.0000 |
| 3 | 45% | 77% | 0.4079 | 0.4207 | 0.9241 | 0.9200 |
| 4 | 13% | 51% | 0.4840 | 0.5479 | 0.5385 | 0.7333 |
| 5 | 32% | 85% | 0.2759 | 0.2969 | 0.9143 | 0.8889 |
| Avg. | – | – | 0.3654 | **0.4006** | 0.8254 | **0.8751** |
| 1 | 1% | 25% | 0.6563 | 0.6875 | 0.5000 | 1.0000 |
| 2 | 5% | 72% | 0.2857 | 0.3000 | 0.6667 | 0.7500 |
| 3 | 45% | 78% | 0.3320 | 0.3561 | 0.8319 | 0.8780 |
| 4 | 14% | 53% | 0.5539 | 0.5637 | 0.9189 | 0.8750 |
| 5 | 35% | 89% | 0.2994 | 0.3307 | 0.8961 | 0.9756 |
| Avg. | – | – | 0.4255 | **0.4476** | 0.7627 | **0.8957** |

Bold values indicate the best results

where $\psi_i$ (Eq. 11) is the weight for participant $i$, computed as the normalized accuracy:

$$\psi_i = \frac{\text{accuracy}_i}{\sum_{j=1}^{n} \text{accuracy}_j}. \tag{11}$$

3. Precision-based federated weighted averaging, with the normalized precision of each model as the weight. Similarly to the previous case, the central server receives both the models and the precision on a test set for each participant, and averages the models parameters with a weighted average where the weights are the normalized precision for each participant. This way, the distributed system amplifies the participant's data with larger precision. The parameters of the federated model $\varpi^*$ are shown in Eq. (12):

$$\varpi^* = \sum_{i=1}^{n} \phi_i \cdot \varpi_i \tag{12}$$

where $\phi_i$ (Eq. 13) is the weight for participant $i$, computed as the normalized accuracy:

Table 19 Blended Blind FL results (constant weights)—Adult dataset

| Agent | Size | % 1s | Accuracy Pre-FL | Accuracy Post-FL | Precision Pre-FL | Precision Post-FL |
|---|---|---|---|---|---|---|
| 1 | 20% | 75% | 0.9166 | 0.9444 | 0.9285 | 0.9285 |
| 2 | 20% | 69% | 0.6944 | 0.9722 | 0.7333 | 0.9615 |
| 3 | 20% | 69% | 0.9444 | 0.9444 | 0.9259 | 0.8620 |
| 4 | 20% | 86% | 0.9444 | 0.9722 | 0.9677 | 1.0000 |
| 5 | 20% | 83% | 0.9722 | 0.9722 | 0.9677 | 0.9677 |
| Avg. | – | – | 0.8944 | **0.9611** | 0.9046 | **0.9439** |
| 1 | 13% | 63% | 0.5000 | 0.5217 | 1.0000 | 1.0000 |
| 2 | 30% | 80% | 0.3043 | 0.3478 | 1.0000 | 1.0000 |
| 3 | 21% | 87% | 0.3077 | 0.3205 | 1.0000 | 1.0000 |
| 4 | 11% | 61% | 0.5581 | 0.6047 | 0.6000 | 1.0000 |
| 5 | 24% | 81% | 0.3137 | 0.3529 | 0.8696 | 0.9005 |
| Avg. | – | – | 0.3968 | **0.4295** | 0.8939 | **0.9810** |
| 1 | 4% | 75% | 0.3125 | 0.3125 | 1.0000 | 0.0000 |
| 2 | 5% | 68% | 0.4210 | 0.4210 | 1.0000 | 0.6666 |
| 3 | 45% | 77% | 0.3374 | 0.3680 | 0.8214 | 0.9285 |
| 4 | 13% | 51% | 0.5217 | 0.5434 | 1.0000 | 1.0000 |
| 5 | 32% | 85% | 0.2991 | 0.3076 | 0.8695 | 0.8000 |
| Avg. | – | – | 0.3783 | **0.3905** | **0.9381** | 0.6790 |
| 1 | 1% | 25% | 0.3333 | 0.3333 | 0.0000 | 0.0000 |
| 2 | 5% | 72% | 0.3500 | 0.3500 | 1.0000 | 1.0000 |
| 3 | 45% | 78% | 0.3105 | 0.3354 | 0.8000 | 0.8461 |
| 4 | 14% | 53% | 0.5102 | 0.5102 | 0.6666 | 0.2500 |
| 5 | 35% | 89% | 0.1904 | 0.1904 | 1.0000 | 1.0000 |
| Avg. | – | – | 0.3389 | **0.3438** | **0.6933** | 0.6192 |

Bold values indicate the best results

$$\phi_i = \frac{\text{precision}_i}{\sum_{j=1}^{n} \text{precision}_j}. \tag{13}$$

## Experimental approach

In all of the following cases we will train two FCM models, using PSO for the optimisation stage, with 20 iterations and a swarm size of 10. The first FCM will have a slope of 2 and use a hyperbolic tangent as activation function, while the second model will have a slope of 5 and a sigmoid activation function.

The first experiment for each dataset will be a baseline model to discuss the case where no distribution is made and all agents build a model as one agent. We will compare these results with the post-federated learning ones to see how our methodology can improve the results of models trained in their individual data (and therefore, models with, in general, worse performance metrics due to the lack of diverse data to be trained with) to obtain similar results to this baseline model.

The other experiments analyse the different combinations of federation methods (Blind Federated Learning and Blended Blind Federated Learning, described in section "Federated learning for FCMs") with the proposed aggregation methods (federated averaging, accuracy-based, and precision-based, defined in section "Aggregation

**Table 20** Blended Blind FL results (constant weights, slope 5, sigmoid)—adult dataset

| Agent | Size | % 1s | Accuracy Pre-FL | Accuracy Post-FL | Precision Pre-FL | Precision Post-FL |
|---|---|---|---|---|---|---|
| 1 | 20% | 75% | 0.9722 | 0.9722 | 1.0000 | 1.0000 |
| 2 | 20% | 69% | 0.8889 | 0.9167 | 0.8750 | 0.9032 |
| 3 | 20% | 69% | 0.9444 | 1.0000 | 1.0000 | 1.0000 |
| 4 | 20% | 86% | 0.9444 | 1.0000 | 0.9231 | 1.0000 |
| 5 | 20% | 83% | 0.9722 | 1.0000 | 0.9630 | 0.9630 |
| Avg. | – | – | 0.9444 | **0.9778** | 0.9522 | **0.9732** |
| 1 | 13% | 63% | 0.3006 | 0.3196 | 0.8929 | 0.9259 |
| 2 | 30% | 80% | 0.3731 | 0.4104 | 0.8261 | 0.8947 |
| 3 | 21% | 87% | 0.3384 | 0.3720 | 0.9167 | 0.8947 |
| 4 | 11% | 61% | 0.3457 | 0.3765 | 0.9565 | 1.0000 |
| 5 | 24% | 81% | 0.3510 | 0.3846 | 0.8391 | 0.8676 |
| Avg. | – | – | 0.3418 | **0.3726** | 0.8862 | **0.9166** |
| 1 | 4% | 75% | 0.3165 | 0.3544 | 0.7333 | 0.6923 |
| 2 | 5% | 68% | 0.2949 | 0.3077 | 0.7500 | 0.6250 |
| 3 | 45% | 77% | 0.3452 | 0.3665 | 0.8837 | 0.8519 |
| 4 | 13% | 51% | 0.4378 | 0.4865 | 0.4286 | 0.6154 |
| 5 | 32% | 85% | 0.2640 | 0.2775 | 0.9077 | 0.9333 |
| Avg. | – | – | 0.3317 | **0.3585** | 0.7407 | **0.7436** |
| 1 | 1% | 25% | 0.5625 | 0.5625 | 0.0000 | 0.0000 |
| 2 | 5% | 72% | 0.3500 | 0.3750 | 0.7143 | 1.0000 |
| 3 | 45% | 78% | 0.3236 | 0.3250 | 0.9506 | 0.8545 |
| 4 | 14% | 53% | 0.4891 | 0.5163 | 0.9231 | 0.7059 |
| 5 | 35% | 89% | 0.2609 | 0.2737 | 0.9130 | 0.9459 |
| Avg. | – | – | 0.3972 | **0.4105** | 0.7002 | **0.7013** |

Bold values indicate the best results

methods"). The authors compare the average accuracy and precision, computed in a test set, for all participants before and after the federation process, and also with the baseline model.

For these experiments, the authors have tested four different data partitions among the participants. The first one is an evenly splitted dataset for every agent. The remaining three are comprised by uneven sets, the first one a random partition and the remaining two with sharp differences where several agents have very small datasets. This way, we can test a hypothetical case where a group of agents want to share secure information and a private model even in the case where one or more of the agents have much less information to share than the rest. Moreover, there are no class balancing mechanisms in the partitioning of the data, and therefore the experiments also test the cases when the percentage of positive samples is noticeable dissimilar.

As it is usual, for each participant's dataset a split train/test has been performed in order to have a validation dataset to compute the performance metrics.

The results will be shown in tables where the rows are the metrics for each participant, and the columns are the following: the size or percentage of the original dataset that each participant has, the percentage of positives in that participant's dataset, and the accuracy and the precision on a test set before and after the Blind Federated Learning process.

**Table 21** Blind FL results (accuracy-based weights)—adult dataset

| Agent | Size | % 1s | Accuracy Pre-FL | Accuracy Post-FL | Precision Pre-FL | Precision Post-FL |
|---|---|---|---|---|---|---|
| 1 | 20% | 69% | 0.9444 | 0.9444 | 0.9600 | 0.9600 |
| 2 | 20% | 78% | 0.8611 | 0.9444 | 0.8965 | 0.9333 |
| 3 | 20% | 86% | 0.8888 | 0.9444 | 0.9354 | 0.9393 |
| 4 | 20% | 78% | 0.9166 | 1.0000 | 0.9032 | 1.0000 |
| 5 | 20% | 69% | 0.9166 | 0.9444 | 1.0000 | 0.9565 |
| Avg. | – | – | 0.9055 | **0.9555** | 0.9390 | **0.9578** |
| 1 | 13% | 63% | 0.4565 | 0.4565 | 1.0000 | 1.0000 |
| 2 | 30% | 80% | 0.3043 | 0.3478 | 1.0000 | 1.0000 |
| 3 | 21% | 87% | 0.2564 | 0.2692 | 0.9231 | 1.0000 |
| 4 | 11% | 61% | 0.5581 | 0.5581 | 1.0000 | 1.0000 |
| 5 | 24% | 81% | 0.2353 | 0.2843 | 0.8462 | 0.8667 |
| Avg. | – | – | 0.3621 | **0.3832** | 0.9538 | **0.9733** |
| 1 | 4% | 75% | 0.3750 | 0.3750 | 1.0000 | 1.0000 |
| 2 | 5% | 68% | 0.3684 | 0.4210 | 1.0000 | 0.5000 |
| 3 | 45% | 77% | 0.3619 | 0.3987 | 0.8055 | 0.8750 |
| 4 | 13% | 51% | 0.5000 | 0.5000 | 0.0000 | 0.0000 |
| 5 | 32% | 85% | 0.2649 | 0.2905 | 0.9333 | 0.8181 |
| Avg. | – | – | 0.3740 | **0.3970** | **0.7478** | 0.6386 |
| 1 | 1% | 25% | 0.3333 | 0.3333 | 0.0000 | 0.0000 |
| 2 | 5% | 72% | 0.3500 | 0.3500 | 1.0000 | 1.0000 |
| 3 | 45% | 78% | 0.2919 | 0.3478 | 0.7727 | 0.9411 |
| 4 | 14% | 53% | 0.5714 | 0.5918 | 0.8571 | 0.6000 |
| 5 | 35% | 89% | 0.2380 | 0.2380 | 1.0000 | 1.0000 |
| Avg. | – | – | 0.3569 | **0.3722** | **0.7259** | 0.7082 |

Bold values indicate the best results

**Experiment 1. Breast Cancer dataset**

In this experiment the authors use the Breast Cancer Wisconsin dataset, made publicly available [42] at the UC Irvine Machine Learning Repository. As a baseline model with no distribution, the FCM with slope 2 and hyperbolic tangent activation function achieves an accuracy on a test set of 0.9211 and a precision of 0.7742, as seen in Table 1, while the FCM with slope 5 and sigmoid function has an accuracy of 0.8246 and a precision of 0.5714, see Table 2.

Our first Blind Federated Learning experiment will consist in a distributed system trained with a methodology that is close to the Blind Federated Learning approach, as described in section "Methodological proposal", and using an aggregation method based on the arithmetic average of the number of participants (federated averaging). Each one of the five participants is provided with a subset of the breast cancer dataset and trains its initial FCM using these data. The results of this first experiment can be found in Tables 3 and 4, where we see that the Blind Federated Learning process improves the values of the accuracy in every case. But for the most uneven data distributions the precision is not increased.

The next experiment uses the Blended Blind Federated Learning process, also described in section 3. As in the previous experiment, the authors will use the arithmetic average of the number of participants as the aggregation method. The results are

**Table 22** Blind FL results (accuracy-based weights, slope 5, sigmoid)—adult dataset

| Agent | Size | % 1s | Accuracy Pre-FL | Accuracy Post-FL | Precision Pre-FL | Precision Post-FL |
|---|---|---|---|---|---|---|
| 1 | 20% | 69% | 0.9722 | 0.9722 | 0.9630 | 0.9630 |
| 2 | 20% | 78% | 0.8889 | 0.9722 | 1.0000 | 1.0000 |
| 3 | 20% | 86% | 0.9444 | 0.9722 | 1.0000 | 1.0000 |
| 4 | 20% | 78% | 0.9444 | 0.9444 | 0.9355 | 0.9355 |
| 5 | 20% | 69% | 0.8889 | 0.9444 | 0.8571 | 0.9231 |
| Avg. | – | – | 0.9278 | **0.9611** | 0.9511 | **0.9643** |
| 1 | 13% | 63% | 0.3834 | 0.4233 | 0.8667 | 0.8846 |
| 2 | 30% | 80% | 0.3140 | 0.3719 | 0.8182 | 0.7857 |
| 3 | 21% | 87% | 0.4182 | 0.4303 | 0.9194 | 0.9020 |
| 4 | 11% | 61% | 0.3462 | 0.3718 | 0.8750 | 0.9286 |
| 5 | 24% | 81% | 0.3175 | 0.3794 | 0.7700 | 0.8434 |
| Avg. | – | – | 0.3559 | **0.3953** | 0.8498 | **0.8688** |
| 1 | 4% | 75% | 0.3023 | 0.3256 | 1.0000 | 1.0000 |
| 2 | 5% | 68% | 0.2319 | 0.3188 | 0.8000 | 1.0000 |
| 3 | 45% | 77% | 0.3191 | 0.3418 | 0.7479 | 0.7705 |
| 4 | 13% | 51% | 0.4681 | 0.4681 | 0.5769 | 0.4500 |
| 5 | 32% | 85% | 0.2766 | 0.2901 | 0.9189 | 0.9057 |
| Avg. | – | – | 0.3740 | **0.3970** | 0.7478 | 0.6386 |
| 1 | 1% | 25% | 0.6667 | 0.6667 | 0.7143 | 0.6667 |
| 2 | 5% | 72% | 0.2949 | 0.3462 | 0.8750 | 1.0000 |
| 3 | 45% | 78% | 0.2889 | 0.3306 | 0.7727 | 0.8415 |
| 4 | 14% | 53% | 0.5829 | 0.6203 | 0.9091 | 0.9615 |
| 5 | 35% | 89% | 0.2893 | 0.3092 | 0.9625 | 0.9710 |
| Avg. | – | – | 0.4245 | **0.4546** | 0.8467 | **0.8881** |

Bold values indicate the best results

described in Tables 5 and 6. As in the previous case, the accuracy increases after the Blind Federated Learning process in all cases, but also the precision is improved for all partitions with this methodology.

It is also worth noticing that, with an even partition, the averaged accuracy of the five models is similar (or even better) than the case of an only model using the full dataset (0.9211 for a unique model vs. 0.9296 for the Federated models in the even case with slope 2 and tangent activation function). Moreover, in the case of the uneven splits, the precision is much better than the baseline model (0.7742 vs. 0.8733 for the first random split).

The next experiment uses the accuracy-based aggregation in order to improve the accuracy values of the model, in combination with the Blind Federated Learning methodology. As previously, there will be four different partitions to understand how this methodology deals with participants with different sizes. Tables 7 and 8 collects the results of this experiment, in which the accuracy improves after the federated learning execution as expected, but we find that the precision levels decrease in all uneven data distributions.

Similarly to the former experiment, the next one uses the accuracy-based aggregation and the four different partitions outlined before, but in this case it will be for a Blended Blind Federated Learning system instead of the conventional one. Tables 9 and 10 show

**Table 23** Blended Blind FL results (accuracy-based weights)—Adult dataset

| Agent | Size | % 1s | Accuracy Pre-FL | Accuracy Post-FL | Precision Pre-FL | Precision Post-FL |
|---|---|---|---|---|---|---|
| 1 | 20% | 75% | 0.9444 | 1.0000 | 1.0000 | 0.9642 |
| 2 | 20% | 94% | 0.9722 | 1.0000 | 1.0000 | 1.0000 |
| 3 | 20% | 78% | 0.9166 | 0.9722 | 0.9629 | 0.9642 |
| 4 | 20% | 78% | 0.9166 | 0.9444 | 0.9310 | 0.9642 |
| 5 | 20% | 78% | 0.8611 | 0.9444 | 0.8484 | 0.9333 |
| Avg. | – | – | 0.9222 | **0.9722** | 0.9484 | **0.9652** |
| 1 | 13% | 63% | 0.4565 | 0.4782 | 1.0000 | 1.0000 |
| 2 | 30% | 80% | 0.3913 | 0.3913 | 1.0000 | 1.0000 |
| 3 | 21% | 87% | 0.2820 | 0.2948 | 1.0000 | 1.0000 |
| 4 | 11% | 61% | 0.5813 | 0.5813 | 0.7500 | 0.7500 |
| 5 | 24% | 81% | 0.2941 | 0.3333 | 1.0000 | 1.0000 |
| Avg. | – | – | 0.4010 | **0.4158** | 0.9500 | 0.9500 |
| 1 | 4% | 75% | 0.4375 | 0.4375 | 1.0000 | 1.0000 |
| 2 | 5% | 68% | 0.3684 | 0.4210 | 0.6666 | 0.6666 |
| 3 | 45% | 77% | 0.3742 | 0.4049 | 0.8157 | 0.8333 |
| 4 | 13% | 51% | 0.5217 | 0.5434 | 0.6666 | 0.6666 |
| 5 | 32% | 85% | 0.2478 | 0.3076 | 0.9230 | 0.8750 |
| Avg. | – | – | 0.3899 | **0.4229** | **0.8144** | 0.8083 |
| 1 | 1% | 25% | 0.5000 | 0.5000 | 0.3333 | 0.0000 |
| 2 | 5% | 72% | 0.3000 | 0.4000 | 1.0000 | 1.0000 |
| 3 | 45% | 78% | 0.3167 | 0.3540 | 0.7857 | 0.8823 |
| 4 | 14% | 53% | 0.4693 | 0.4897 | 0.5000 | 0.5000 |
| 5 | 35% | 89% | 0.1587 | 0.1746 | 0.8750 | 1.0000 |
| Avg. | – | – | 0.3490 | **0.3837** | **0.6988** | 0.6765 |

Bold values indicate the best results

how the Blended Blind Federated Learning methodology improves the results of the previous experiment, in the sense that not only the accuracy is better after the federated learning execution, but also the precision in all cases but one.

Our next set of experiments will deal with the precision-based aggregation method in order to try to improve the precision of the model, since the accuracy improves in every previous test. The results for the Blind Federated Learning methodology, using the four different partitions as previously, can be found in Table 11 and 12, and show the usual improvement of the accuracy of the model post federated learning, but also an increase in the precision of the model in all cases but one.

Finally, the last experiment will be similar to the previous one: the Blended Blind Federated Learning approach, with the precision-based aggregation method, and with the usual four different partitions. The results are described in Tables 13 and 14 and show that the accuracy keeps improving even when the weights of the aggregation method depend on the precision, but the precision only increases in two out of four cases.

**Experiment 2. Adult dataset**

The dataset for the second experiment will be the adult dataset, with census data from 1994 containing demographic features of adults and their income, from the US Census Bureau, and publicly available [42] at the UC Irvine ML Repository. The baseline

**Table 24** Blended Blind FL results (accuracy-based weights, slope 5, sigmoid)—Adult dataset

| Agent | Size | % 1s | Accuracy Pre-FL | Accuracy Post-FL | Precision Pre-FL | Precision Post-FL |
|---|---|---|---|---|---|---|
| 1 | 20% | 75% | 0.8889 | 0.9444 | 0.9565 | 0.9231 |
| 2 | 20% | 94% | 0.9444 | 1.0000 | 1.0000 | 1.0000 |
| 3 | 20% | 78% | 0.8333 | 0.8889 | 0.8800 | 0.8276 |
| 4 | 20% | 78% | 1.0000 | 1.0000 | 1.0000 | 0.9600 |
| 5 | 20% | 78% | 0.7500 | 1.0000 | 0.7188 | 1.0000 |
| Avg. | – | – | 0.8833 | **0.9667** | 0.9111 | **0.9421** |
| 1 | 13% | 63% | 0.3292 | 0.3727 | 0.8485 | 0.9091 |
| 2 | 30% | 80% | 0.3857 | 0.4000 | 0.9200 | 0.9091 |
| 3 | 21% | 87% | 0.3539 | 0.3896 | 0.8837 | 0.8929 |
| 4 | 11% | 61% | 0.3663 | 0.3779 | 0.9231 | 0.8636 |
| 5 | 24% | 81% | 0.3205 | 0.3413 | 0.9333 | 0.8594 |
| Avg. | – | – | 0.3511 | **0.3763** | **0.9017** | 0.8868 |
| 1 | 4% | 75% | 0.3171 | 0.3780 | 0.6667 | 0.7692 |
| 2 | 5% | 68% | 0.3553 | 0.3684 | 0.9231 | 1.0000 |
| 3 | 45% | 77% | 0.3980 | 0.4162 | 0.9242 | 0.9259 |
| 4 | 13% | 51% | 0.5106 | 0.5426 | 0.8667 | 0.6667 |
| 5 | 32% | 85% | 0.2659 | 0.2897 | 0.9206 | 0.9184 |
| Avg. | – | – | 0.3694 | **0.3990** | **0.8603** | 0.8560 |
| 1 | 1% | 25% | 0.5926 | 0.5926 | 0.5000 | 0.5000 |
| 2 | 5% | 72% | 0.2987 | 0.3377 | 0.6429 | 0.6667 |
| 3 | 45% | 78% | 0.3205 | 0.3438 | 0.7900 | 0.8163 |
| 4 | 14% | 53% | 0.5213 | 0.5403 | 0.9565 | 0.9167 |
| 5 | 35% | 89% | 0.2538 | 0.2885 | 0.9000 | 0.9706 |
| Avg. | – | – | 0.3974 | **0.4206** | 0.7579 | **0.7740** |

Bold values indicate the best results

FCM model (Table 15) reaches an accuracy of 0.8611 and a precision of 0.9487 in the test set for the FCM model with slope 2 and hyperbolic tangent activation function, and an accuracy of 0.9222 and a precision of 0.9257 (Table 16).

The Blind Federated Learning approach combined with arithmetic average of the number of participants in the second dataset shows a very different behaviour for the even and uneven splits (Table 17 and 18): the federated learning process improves the performance metrics for the even split compared with the model without federation, but for all the uneven splits we see that the accuracy is much lower because of the amount of data and the target imbalance. Nevertheless, the federation process improves the accuracy, but not the precision, which has higher values than the accuracy.

Next, the Blended Blind Federated Learning with constant weights shows (Tables 19 and 20) similar results to those of the Blind Federated Learning, with high accuracy and precision for the even split and lower accuracy for the uneven, and improvement after the federation in the accuracy but not for the precision.

The results (Tables 21 and 22) for the Blind Federated Learning using an aggregation based in accuracy shows similar results to the two previous experiments, with limited accuracy for uneven splits and no improvements in the precision for the most extreme uneven splits.

**Table 25** Blind FL results (Precision-based weights)—Adult dataset

| Agent | Size | % 1s | Accuracy Pre-FL | Accuracy Post-FL | Precision Pre-FL | Precision Post-FL |
|---|---|---|---|---|---|---|
| 1 | 20% | 67% | 0.9444 | 0.9444 | 0.9230 | 0.9166 |
| 2 | 20% | 61% | 0.8611 | 0.8888 | 0.8400 | 0.7777 |
| 3 | 20% | 72% | 0.8055 | 0.8888 | 0.7878 | 0.8666 |
| 4 | 20% | 78% | 0.9722 | 0.9722 | 1.0000 | 0.9642 |
| 5 | 20% | 72% | 0.8611 | 0.9444 | 0.9200 | 0.9259 |
| Avg. | – | – | 0.8889 | **0.9278** | **0.8941** | 0.8902 |
| 1 | 13% | 63% | 0.5217 | 0.5217 | 1.0000 | 1.0000 |
| 2 | 30% | 80% | 0.3260 | 0.3478 | 0.9231 | 1.0000 |
| 3 | 21% | 87% | 0.2179 | 0.2948 | 1.0000 | 1.0000 |
| 4 | 11% | 61% | 0.6046 | 0.6046 | 1.0000 | 1.0000 |
| 5 | 24% | 81% | 0.3823 | 0.3921 | 0.9000 | 0.8947 |
| Avg. | – | – | 0.4106 | **0.4322** | 0.9646 | **0.9789** |
| 1 | 4% | 75% | 0.3125 | 0.3125 | 1.0000 | 1.0000 |
| 2 | 5% | 68% | 0.3157 | 0.3684 | 0.5000 | 1.0000 |
| 3 | 45% | 77% | 0.4171 | 0.4355 | 0.9428 | 0.8750 |
| 4 | 13% | 51% | 0.5434 | 0.6086 | 1.0000 | 1.0000 |
| 5 | 32% | 85% | 0.2905 | 0.3418 | 0.8636 | 0.8500 |
| Avg. | – | – | 0.3759 | **0.4134** | 0.8612 | **0.9450** |
| 1 | 1% | 25% | 0.1666 | 0.3333 | 0.0000 | 0.2500 |
| 2 | 5% | 72% | 0.3500 | 0.4000 | 1.0000 | 1.0000 |
| 3 | 45% | 78% | 0.3416 | 0.3726 | 0.9166 | 0.8947 |
| 4 | 14% | 53% | 0.5510 | 0.5918 | 0.7000 | 0.5000 |
| 5 | 35% | 89% | 0.2222 | 0.2301 | 1.0000 | 1.0000 |
| Avg. | – | – | 0.3263 | **0.3855** | 0.7233 | **0.7289** |

Bold values indicate the best results

Tables 23 and 24 show the results for the Blended Blind Federated Learning with accuracy-based weights, very similar to those with Blind Federated Learning and the same aggregation method.

The precision-based aggregation method should drastically improve the precision, a metric that, as we have seen in the previous experiments, is more prone to not being improved by the federation process. This is the behaviour that we see in the experiments with this aggregation method, as shown by Table 25 and 26, where not only the accuracy improves for all splits, but also the precision for all but one, the even split.

In the case of the Blended Blind Federated Learning with precision-based aggregation (Tables 27 and 28), the accuracy is improved in all cases, and the precision is also improved but only in the first case of the even split.

### Discussion

The different experiments show that, even in the most imbalanced cases, the federated learning approach improves the average accuracy of the models. FL increases the performance of the models while allowing the private sharing of data among the participants.

Nevertheless, there are differences between the efficiency of the models after the distinct federation methodologies. In general terms, the Blind Federated Learning method has lower precision than the Blended counterpart with the same aggregation

**Table 26** Blind FL results (Precision-based weights, slope 5, sigmoid)—adult dataset

| Agent | Size | % 1s | Accuracy Pre-FL | Accuracy Post-FL | Precision Pre-FL | Precision Post-FL |
|---|---|---|---|---|---|---|
| 1 | 20% | 67% | 0.9167 | 0.9722 | 0.9630 | 0.9655 |
| 2 | 20% | 61% | 0.9722 | 1.0000 | 1.0000 | 0.9231 |
| 3 | 20% | 72% | 0.7500 | 0.9722 | 0.7500 | 0.9643 |
| 4 | 20% | 78% | 0.8889 | 0.8889 | 0.9000 | 0.9000 |
| 5 | 20% | 72% | 0.9167 | 0.9444 | 0.9231 | 0.8621 |
| Avg. | – | – | 0.8889 | **0.9556** | 0.9072 | **0.9230** |
| 1 | 13% | 63% | 0.3201 | 0.3537 | 0.8718 | 0.8333 |
| 2 | 30% | 80% | 0.3621 | 0.3793 | 0.8500 | 0.7143 |
| 3 | 21% | 87% | 0.3692 | 0.3754 | 0.9623 | 0.9730 |
| 4 | 11% | 61% | 0.4740 | 0.4805 | 0.9722 | 0.9259 |
| 5 | 24% | 81% | 0.3255 | 0.3458 | 0.8632 | 0.9111 |
| Avg. | – | – | 0.3702 | **0.3869** | 0.9039 | 0.8715 |
| 1 | 4% | 75% | 0.3117 | 0.3247 | 0.8182 | 0.7000 |
| 2 | 5% | 68% | 0.2821 | 0.2821 | 1.0000 | 0.6667 |
| 3 | 45% | 77% | 0.3495 | 0.3774 | 0.8532 | 0.8514 |
| 4 | 13% | 51% | 0.5161 | 0.5215 | 0.8421 | 0.6429 |
| 5 | 32% | 85% | 0.2851 | 0.3050 | 0.8750 | 0.8980 |
| Avg. | – | – | 0.3489 | **0.3621** | **0.8777** | 0.7518 |
| 1 | 1% | 25% | 0.5588 | 0.5882 | 0.3333 | 0.0000 |
| 2 | 5% | 72% | 0.2800 | 0.2800 | 1.0000 | 0.8000 |
| 3 | 45% | 78% | 0.3645 | 0.3862 | 0.8915 | 0.9506 |
| 4 | 14% | 53% | 0.4754 | 0.4809 | 0.7500 | 0.4667 |
| 5 | 35% | 89% | 0.2393 | 0.2935 | 0.8025 | 0.9028 |
| Avg. | – | – | 0.3836 | **0.4057** | **0.7555** | 0.6240 |

Bold values indicate the best results

approach in two out of three datasets used, while in the third one the results are quite similar for all aggregations. For the breast cancer dataset the results show that, in the case of the Blind Federated Learning, the averaged precision of the models does not improve after the federation process in two out of four examples with different imbalanced data (see Tables 3 and 4), while in all cases of the new methodology the averaged precision increases (see Tables 5 and 6).

A similar reasoning can be applied to the accuracy-based aggregation methods. In this case, as mentioned before, the accuracy improves in all cases after the federated learning process, independently of the methodology used. Nevertheless, as in the previous example, the Blind Federation process does not increase the model precision in three out of four experiments, always in the imbalanced cases (see Table 7). Meanwhile, for the Blended Federated Learning procedure, the precision is increased in all cases but one, the second most imbalanced one (see Table 9).

Given the difficulties shown to improve the average precision of the models, the authors test if the precision-based aggregation method can improve the precision of the models while maintaining the accuracy levels post-Federation. The results provides two insights. Firstly, the performance of the model regarding accuracy is improved in all cases, just like in the previous experiments. Secondly the precision is improved in most of the experiments with this aggregation method.

**Table 27** Blended Blind FL results (Precision-based weights)—adult dataset

| Agent | Size | % 1s | Accuracy Pre-FL | Accuracy Post-FL | Precision Pre-FL | Precision Post-FL |
|---|---|---|---|---|---|---|
| 1 | 20% | 72% | 0.7777 | 0.8888 | 0.8214 | 0.8928 |
| 2 | 20% | 72% | 0.9166 | 0.9444 | 0.9600 | 1.0000 |
| 3 | 20% | 81% | 0.8888 | 0.9444 | 0.9629 | 1.0000 |
| 4 | 20% | 69% | 0.8611 | 0.9166 | 0.9166 | 0.9200 |
| 5 | 20% | 83% | 0.9166 | 0.9166 | 0.9354 | 0.9354 |
| Avg. | – | – | 0.8722 | **0.9222** | 0.9193 | **0.9497** |
| 1 | 13% | 63% | 0.4680 | 0.5106 | 0.5710 | 0.6667 |
| 2 | 30% | 80% | 0.3296 | 0.3296 | 1.0000 | 1.0000 |
| 3 | 21% | 87% | 0.2631 | 0.2894 | 1.0000 | 1.0000 |
| 4 | 11% | 61% | 0.5454 | 0.6136 | 0.6667 | 0.6000 |
| 5 | 24% | 81% | 0.3619 | 0.3619 | 1.0000 | 0.8000 |
| Avg. | – | – | 0.3936 | **0.4210** | **0.8476** | 0.8133 |
| 1 | 4% | 75% | 0.3888 | 0.3888 | 1.0000 | 1.0000 |
| 2 | 5% | 68% | 0.2380 | 0.2857 | 0.5000 | 0.0000 |
| 3 | 45% | 77% | 0.3293 | 0.3652 | 1.0000 | 0.7222 |
| 4 | 13% | 51% | 0.4772 | 0.5454 | 0.0000 | 0.5000 |
| 5 | 32% | 85% | 0.3214 | 0.3482 | 0.9333 | 0.8333 |
| Avg. | – | – | 0.3510 | **0.3867** | **0.6867** | 0.6111 |
| 1 | 1% | 25% | 0.6250 | 0.6250 | 0.0000 | 0.0000 |
| 2 | 5% | 72% | 0.4117 | 0.4705 | 0.7142 | 0.8000 |
| 3 | 45% | 78% | 0.3488 | 0.3604 | 0.8437 | 0.7222 |
| 4 | 14% | 53% | 0.5531 | 0.5531 | 0.6666 | 0.5000 |
| 5 | 35% | 89% | 0.1949 | 0.2372 | 1.0000 | 0.8750 |
| Avg. | – | – | 0.4267 | **0.4493** | **0.6449** | 0.5794 |

Bold values indicate the best results

The Blind Federated Learning process increases the averaged precision of the models in all cases but one, the most imbalanced test (see Tables 11 and 12). Despite of the rest of the experiments using different aggregation methods, the Blind methodology performs better than the Blended one, since the averaged precision is only increased in two out of four tests (see Tables 13 and 14).

Finally, the authors benchmarked the different methodologies and aggregation methods with the initial case where no distribution is made, to understand if the accuracy and precision metrics of the distributed problems are similar to a conventional problem.

The performance of an FCM with only one agent and trained using the full dataset is shown in Tables 1, 2, 15 and 16. The accuracy and precision are similar to the averaged metrics of the Federated models using the new methodology, with all aggregation methods, in the case of balanced datasets.

In the case of the Blind Federated Learning approach, we can see that the absolute performance values for the balanced datasets are worse than using the Blended Blind Federated Learning procedure: 0.9123 for the averaged accuracy and 0.6981 for the averaged precision with the aggregation using constant weights, 0.9209 for the

**Table 28** Blended Blind FL results (Precision-based weights, slope 5, sigmoid)—Adult dataset

| Agent | Size | % 1s | Accuracy Pre-FL | Accuracy Post-FL | Precision Pre-FL | Precision Post-FL |
|---|---|---|---|---|---|---|
| 1 | 20% | 72% | 0.8889 | 0.9444 | 0.8966 | 0.9000 |
| 2 | 20% | 72% | 0.8889 | 1.0000 | 0.9630 | 1.0000 |
| 3 | 20% | 81% | 0.9722 | 0.9722 | 1.0000 | 1.0000 |
| 4 | 20% | 69% | 1.0000 | 1.0000 | 1.0000 | 0.9688 |
| 5 | 20% | 83% | 0.9167 | 0.9722 | 0.9000 | 0.9643 |
| Avg. | – | – | 0.9333 | **0.9778** | 0.9519 | **0.9666** |
| 1 | 13% | 63% | 0.3007 | 0.3277 | 0.7692 | 0.8205 |
| 2 | 30% | 80% | 0.3651 | 0.3730 | 0.8696 | 0.8182 |
| 3 | 21% | 87% | 0.2708 | 0.3095 | 0.7045 | 0.7500 |
| 4 | 11% | 61% | 0.3618 | 0.3750 | 0.9524 | 1.0000 |
| 5 | 24% | 81% | 0.3115 | 0.3496 | 0.8429 | 0.9063 |
| Avg. | – | – | 0.3220 | **0.3470** | 0.8277 | **0.8590** |
| 1 | 4% | 75% | 0.2785 | 0.3038 | 0.6667 | 0.6000 |
| 2 | 5% | 68% | 0.3200 | 0.4000 | 0.7778 | 1.0000 |
| 3 | 45% | 77% | 0.3324 | 0.3709 | 0.9157 | 0.9740 |
| 4 | 13% | 51% | 0.4433 | 0.4588 | 0.7500 | 0.7500 |
| 5 | 32% | 85% | 0.3062 | 0.3159 | 0.9722 | 0.9773 |
| Avg. | – | – | 0.3361 | **0.3699** | 0.8165 | **0.8603** |
| 1 | 1% | 25% | 0.6875 | 0.7188 | 0.6667 | 0.5000 |
| 2 | 5% | 72% | 0.2692 | 0.2949 | 0.8182 | 0.7143 |
| 3 | 45% | 78% | 0.3045 | 0.3361 | 0.7232 | 0.7973 |
| 4 | 14% | 53% | 0.5000 | 0.5412 | 0.7419 | 0.7368 |
| 5 | 35% | 89% | 0.3145 | 0.3183 | 0.9271 | 0.8889 |
| Avg. | – | – | 0.4152 | **0.4418** | **0.7754** | 0.7275 |

Bold values indicate the best results

averaged accuracy and 0.6981 for the averaged precision with the accuracy-based aggregation, and 0.9123 for the averaged accuracy and 0.7137 for the averaged precision with the precision-based aggregation, in the case of the FCM with slope 2 and hyperbolic tangent activation function.

Clearly, the performance metrics for the imbalanced cases are non-comparable to the non-distributed problem, since the difference between the amount of information each participant holds has to be leveraged.

## Conclusions

In this research, the authors propose two innovative methodologies to apply federated learning to FCMs, in order to take advantage of the benefits of this new paradigm for Distributed Artificial Intelligence that allows the sharing of private data in a secure way to train a sophisticated machine learning model.

Both methods show an improvement of the averaged accuracy post-Federation in all experiments performed, both in balanced and imbalanced data. In the balanced case, we can see that both the accuracy and precision are comparable to the performance metrics of the non-distributed case. That is, an only FCM trained with all the available data.

Finally, comparing the two presented proposals, the Blind Federated Learning, and Blended Blind Federated Learning, the second proposal, where the new local model is obtained by averaging the parameters of the global method with the ones of the previous local method, instead of using the global model as the new local model as in the case of the Blind Federated Learning approach, generally performs better across all experiments but for the case when a precision-based aggregation is used.

Also, an important benefit to the use of FCM for the federated learning approach is that there is no need for the definition of an initial model as in the case of conventional federated learning with neural networks, where an additional central server usually describes the architecture of the network that every participant will train. In this case, every participant trains the FCM without any predefined model from the server, making it a blind method. The presented approach addresses the challenge of federated learning without the requirement of an initial model, constituting a novelty in the field.

In the course of this research, we have proposed a methodology for federated learning without an initial model, primarily relying on FCMs. While this research has addressed the challenges associated with blind federated learning, an open research question remains unexplored: the development of federated models based on AI architectures different from FCMs. When dealing with AI models other than FCMs, the blind approach needs to be reformulated, as these models lack the specific characteristics of FCMs.

**Acknowledgements**
Prof. Salmeron research was kindly supported by the project Artificial Intelligence for Healthy Aging (Convocatoria 2021 – Misiones de I+D en Inteligencia Artificial: Inteligencia Artificial distribuida para el diagnóstico y tratamiento temprano de enfermedades con gran prevalencia en el envejecimiento, exp.: MIA.2021.M02.0007) lead by Capgemini Engineering.

**Author contributions**
All authors collaborated equally to the final manuscript.

**Funding**
The authors declare that they have no funding to disclose.

**Availability of data and materials**
The datasets analysed during the current study are available in the UCI repository, https://archive.ics.uci.edu/ml/datasets.php.

**Declarations**

**Ethics approval and consent to participate**
Not applicable.

**Consent for publication**
Not applicable.

**Competing interests**
The authors declare that they have no competing interests.

Received: 12 March 2023   Accepted: 7 April 2024
Published online: 23 April 2024

## Publisher's Note

Springer Nature remains neutral with regard to jurisdictional claims in published maps and institutional affiliations.